\documentclass[final]{cvpr}

\usepackage{times}
\usepackage{epsfig}
\usepackage{graphicx}
\usepackage{amsmath}
\usepackage{amssymb}

\usepackage[pagebackref=true,breaklinks=true,colorlinks,bookmarks=false]{hyperref}

\usepackage{comment}
\usepackage{amsmath,amssymb} 
\usepackage{color}
\usepackage{booktabs}
\usepackage{multirow}
\usepackage{multicol}
\usepackage{xspace}
\usepackage{xcolor}
\usepackage{adjustbox}
\usepackage[switch]{lineno}
\usepackage{threeparttable}
\usepackage{import}
\usepackage{subcaption}
\usepackage{nopageno}

\def\SSV2{\textit{SSV2}\xspace}
\def\mSSV2{\textit{Mini-SSV2}\xspace}
\def\Kinetics{\textit{Kinetics}\xspace}
\def\mKinetics{\textit{Mini-Kinetics}\xspace}
\def\Moments{\textit{MiT}\xspace}
\def\mMoments{\textit{Mini-MiT}\xspace}

\newcommand{\mycomment}[1]{}

\def\cvprPaperID{3850} 
\def\confYear{CVPR 2021}
\graphicspath{{Main/},{Supplemental/}}

\begin{document}


\title{Deep Analysis of CNN-based Spatio-temporal Representations for\\ Action Recognition}

\author{Chun-Fu (Richard) Chen$^{1,\dagger}$, Rameswar Panda$^{1,\dagger}$, Kandan Ramakrishnan$^{1}$,\\ Rogerio Feris$^1$, John Cohn$^1$, Aude Oliva$^2$, Quanfu Fan$^{1,\dagger}$\\
$\dagger$: Equal Contribution \\
$^1$MIT-IBM Watson AI Lab, $^2$Massachusetts Institute of Technology
}

\maketitle

\begin{abstract}
In recent years, a number of approaches based on 2D or 3D convolutional neural networks (CNN) have emerged for video action recognition, achieving state-of-the-art results on several large-scale benchmark datasets. In this paper, we carry out in-depth comparative analysis to better understand the differences between these approaches and the progress made by them. To this end, we develop an unified framework for both 2D-CNN and 3D-CNN action models, which enables us to remove bells and whistles and provides a common ground for fair comparison. We then conduct an effort towards a large-scale analysis involving over 300 action recognition models. Our comprehensive analysis reveals that a) a significant leap is made in efficiency for action recognition, but not in accuracy; b) 2D-CNN and 3D-CNN models behave similarly in terms of spatio-temporal representation abilities and transferability. Our codes are available at \url{https://github.com/IBM/action-recognition-pytorch}. 

\end{abstract}
\section{Introduction}
\label{sec:intro}


With the recent advances in convolutional neural networks (CNNs)~\cite{InceptionV1:Szegedy_2015_CVPR,He_2016_CVPR} and the availability of large-scale video benchmark datasets~\cite{Kinetics:kay2017kinetics,Moments:monfort2019moments}, deep learning approaches have dominated the field of video action recognition
by using 2D-CNNs~\cite{TSN:wang2016temporal,TSM:lin2018temporal,bLVNetTAM} or 3D-CNNs~\cite{I3D:carreira2017quo,ResNet3D_2:hara2018can, SlowFast:feichtenhofer2018slowfast} or both ~\cite{luo2019grouped, sudhakaran2020gate}. The 2D CNNs perform temporal modeling independent of 2D spatial convolutions while their 3D counterparts learn space and time information jointly by 3D convolution. These methods have achieved state-of-the-art performance on multiple large-scale benchmarks such as Kinetics~\cite{Kinetics:kay2017kinetics} and Something-Something~\cite{Something:goyal2017something}.

\begin{figure}[bth!]
    \centering
    \includegraphics[width=0.85\linewidth]{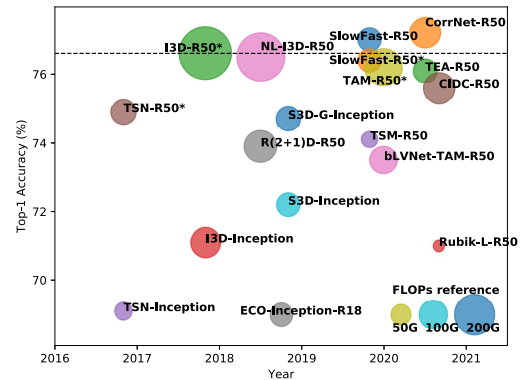} \vspace{-3mm}
    \caption{\small Recent progress of action recognition on Kinetics-400 (only models based on InceptionV1 and ResNet50 are included). Models marked with $*$ are re-trained and evaluated (see Section~\ref{subsec:SOTA}) while others are from the existing literature. The size of a circle indicates the 1-clip FLOPs of a model. With temporal pooling turned off, I3D performs on par with the state-of-the-art approaches. Best viewed in color.}
    \label{fig:perf-comparison} 
    \vspace{-10pt}
\end{figure}

Although CNN-based approaches have made impressive progress in action recognition, there are several fundamental questions that still largely remain unanswered in the field. For example, what contributes to improved spatio-temporal representations of these recent approaches? Do these approaches enable more effective temporal modeling, the crux of the matter for action recognition? Furthermore, there seems no clear winner between 2D-CNN and 3D-CNN approaches in terms of accuracy. 3D models report better performance than 2D models on Kinetics while the latter are superior on Something-Something. How differently do these two types of models behave with regard to spatial-temporal modeling of video data? 

We argue that the difficulty of understanding the recent progress on action recognition is mainly due to the lack of fairness in performance evaluation related to datasets, backbones and experimental practices. 
In contrast to image recognition where ImageNet~\cite{deng2009imagenet} has served as a gold-standard benchmark for evaluation, there are at least 4$\sim$5 popular action datasets widely used for evaluation (see Figure~\ref{fig:challenge}). While Kinetics-400~\cite{Kinetics:kay2017kinetics} has recently emerged as a primary benchmark for action recognition, it is known to be strongly biased towards spatial modeling, thus being inappropriate for validating a model's capability of spatio-temporal modeling. In addition, there seems to be a tendency in current research to overly focus on pursuing state-of-the-art (SOTA) performance, but overlooking other important factors such as the backbone networks and the number of input frames. For instance,
I3D~\cite{I3D:carreira2017quo} based on 3D-InceptionV1 has become a \textit{``gatekeeper''} baseline to compare with for any recently proposed approaches of action recognition. However such comparisons are often unfair against stronger backbones such as ResNet50~\cite{He_2016_CVPR}. As shown in Figure~\ref{fig:perf-comparison}, I3D, with ResNet50 as backbone, performs comparably with or outperforms many recent methods that are claimed to be better. As a result, such evaluation are barely informative w.r.t whether the improved results of an approach come from a better backbone or the algorithm itself. As discussed in Section~\ref{sec:eval-challenge}, performance evaluation in action recognition may be further confounded by many other issues such as variations in training and evaluation protocols, model inputs and pretrained models.

In light of the great need for 
better understanding of CNN-based action recognition models, in this paper we provide a common ground for comparative analysis of 2D-CNN and 3D-CNN models without any bells and whistles. 
We conduct comprehensive experiments and analysis to compare several representative 2D-CNN and 3D-CNN methods on three large-scale benchmark datasets. Our main goal is to deliver deep understanding of the important questions brought up above, especially,  \emph{a) the current progress of action recognition and b) the differences between 2D-CNN and 3D-CNN methods w.r.t spatial-temporal representations of video data.} 
Our systematic analysis provides insights to researchers to understand spatio-temporal effects of different action models across backbone and architecture and will broadly simulate discussions in the community regarding a very important but largely neglected issue of fair comparison in video action recognition.

The main contributions of our work as follows:

\begin{itemize}
\itemsep-0.1em 
\item \textbf{A unified framework for Action Recognition.} We present a unified framework for 2D-CNN and 3D-CNN approaches and implement several representative methods for comparative analysis on three standard action recognition benchmark datasets.
\item \textbf{Spatio-Temporal Analysis.} We systematically compare 2D-CNN and 3D-CNN models to better understand the differences and spatio-temporal behavior of these models. Our analysis leads to some interesting findings as follows: a) Temporal  pooling  tends  to  suppress  the  efficacy  of temporal modeling in an action model, but surprisingly provides a significant performance boost to TSN~\cite{TSN:wang2016temporal}; b) By removing non-structural differences between 2D-CNN and 3D-CNN models, they behave similarly in terms of spatio-temporal representation abilities and transferability.


\item \textbf{Benchmarking of SOTA Approaches.} We thoroughly benchmarked several SOTA approaches and compared them with I3D. Our analysis reveals that I3D still stays on par with SOTA approaches in terms of accuracy (Figure~\ref{fig:perf-comparison}) and the recent advance in action recognition is mostly on the efficiency side, not on accuracy. Our analysis also suggests that the input sampling strategy taken by a model (i.e. uniform or dense sampling) should be considered for fairness when comparing two models (Section~\ref{subsec:SOTA}).


\end{itemize}

\section{Related Work}
\label{sec:literature}

Video understanding has made rapid progress with the introduction of a number of large-scale video datasets such as Kinetics~\cite{Kinetics:kay2017kinetics}, Sports1M~\cite{karpathy2014large}, Moments-In-Time~\cite{Moments:monfort2019moments}, and YouTube-8M~\cite{abu2016youtube}. A number of models introduced recently have emphasized the need to efficiently model spatio-temporal information for video action recognition. 

Most successful deep architectures for action recognition are usually based on two-stream model~\cite{Simonyan14TwoStream}, processing RGB frames and optical-flow in two separate CNNs with a late fusion in upper layers~\cite{karpathy2014large}. 
Two-stream approaches have been used in different action recognition methods~\cite{cheron2015p,lstm:donahue2015longterm,gkioxari2015finding,yue2015beyond,srivastava2015unsupervised,venugopalan2015sequence,weinzaepfel2015learning,wang2015action,feichtenhofer2016spatiotemporal,feichtenhofer2017spatiotemporal}.
Another straightforward but popular approach is the use of 2D-CNN to extract frame-level features and then model the temporal causality. For example, TSN~\cite{TSN:wang2016temporal} propose consensus module to aggregate features; on the other hand,  TRN~\cite{TRN:zhou2018temporal} use bag of features to model relationship between frames. 
While TSM~\cite{TSM:lin2018temporal} shifts part of the channels along temporal dimension, thereby allowing for information to be exchanged among neighboring frames, TAM~\cite{bLVNetTAM} is based on depthwise $1\times1$ convolutions to capture temporal dependencies across frames effectively. Different methods for temporal aggregation of feature descriptors have also been proposed ~\cite{fernando2015modeling,lev2016rnn,xu2015discriminative,wang2015action,peng2014action,girdhar2017actionvlad,girdhar2019video}. More complex approaches have also been investigated for capturing long-range dependencies, e.g. non-local neural networks~\cite{Nonlocal:Wang2018NonLocal}. 


Another approach is to use 3D-CNN, which extends the success of 2D models in image recognition~\cite{3DCNNHuamn:Ji2013} to recognize actions in videos. For example, C3D~\cite{C3D:Tran2015learning} learns 3D ConvNets which outperforms 2D CNNs through the use of large-scale video datasets. 
Many variants of 3D-CNNs are introduced for learning spatio-temporal features such as I3D~\cite{I3D:carreira2017quo} and ResNet3D~\cite{ResNet3D_2:hara2018can}. 3D CNN features were also demonstrated to generalize well to other vision tasks, such as action detection~\cite{shou2016temporal}, video captioning~\cite{pan2016jointly}, action localization~\cite{paul2018w}, and  video summarization~\cite{panda2017collaborative}. Nonetheless, as 3D convolution leads high computational load, few works aim to reduce the complexity by decomposing the 3D convolution into 2D spatial convolution and 1D temporal convolution, e.g., P3D~\cite{P3D:Qiu_2017_ICCV}, S3D~\cite{S3D}, R(2+1)D~\cite{R2plus1D:Tran_2018_CVPR}, or incorporating group convolution~\cite{CSN:Tran_2019_ICCV}; or using a combination of 2D-CNN and 3D-CNN~\cite{ECO:zolfaghari2018eco}. Furthermore, SlowFast network employs two pathways to capture short-term and long-term temporal information~\cite{SlowFast:feichtenhofer2018slowfast} by processing a video at both slow and fast frame rates. Beyond that, Timeception applies the Inception concept in the temporal domain for capturing long-range temporal dependencies~\cite{Timeception:Hussein_CVPR_2019}. Feichtenhofer~\cite{X3D_Feichtenhofer_2020_CVPR} finds efficient networks by extending 2D architectures through a stepwise expansion approach over the key variables such as temporal duration, frame rate, spatial resolution, network width, etc. Leveraging weak supervision~\cite{ghadiyaram2019large, wang2017untrimmednets,kuehne2017weakly} or distillation~\cite{girdhar2019distinit} is another recent trend in action recognition.  
Few works have assessed the importance of temporal information in a video, e.g., Sigurdsson \emph{et.al} analyzed performance per action category based on different levels of object complexity, verb complexity, and motion~\cite{sigurdsson2017actions}. They state that to differentiate temporally similar but semantically different videos, its important for models to develop temporal understanding. Huang~\emph{et. al} analyzed the effect of motion via an ablation analysis on C3D model~\cite{huang2018makes}. Nonetheless, these works only study a limited set of backbone and temporal modeling methods. 

\section{Challenges of Evaluating Action Models}
\label{sec:eval-challenge}

\begin{figure}[t]
    \centering
    \begin{subfigure}{0.6\linewidth}
        \centering
      \includegraphics[width=1\linewidth]{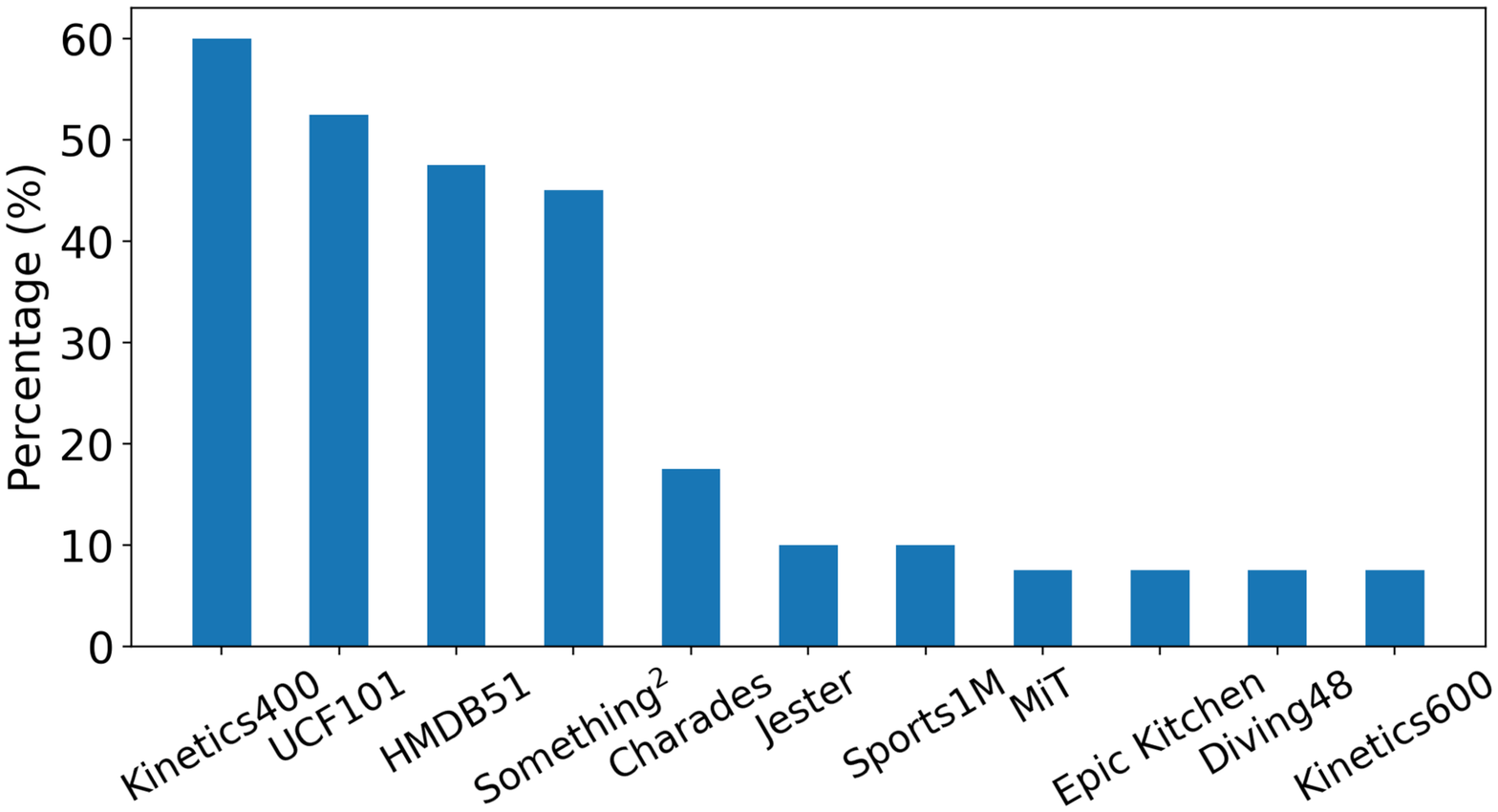}
    \end{subfigure}
    \hfill
    \begin{subfigure}{0.35\linewidth}
    \centering
      \includegraphics[width=1\linewidth]{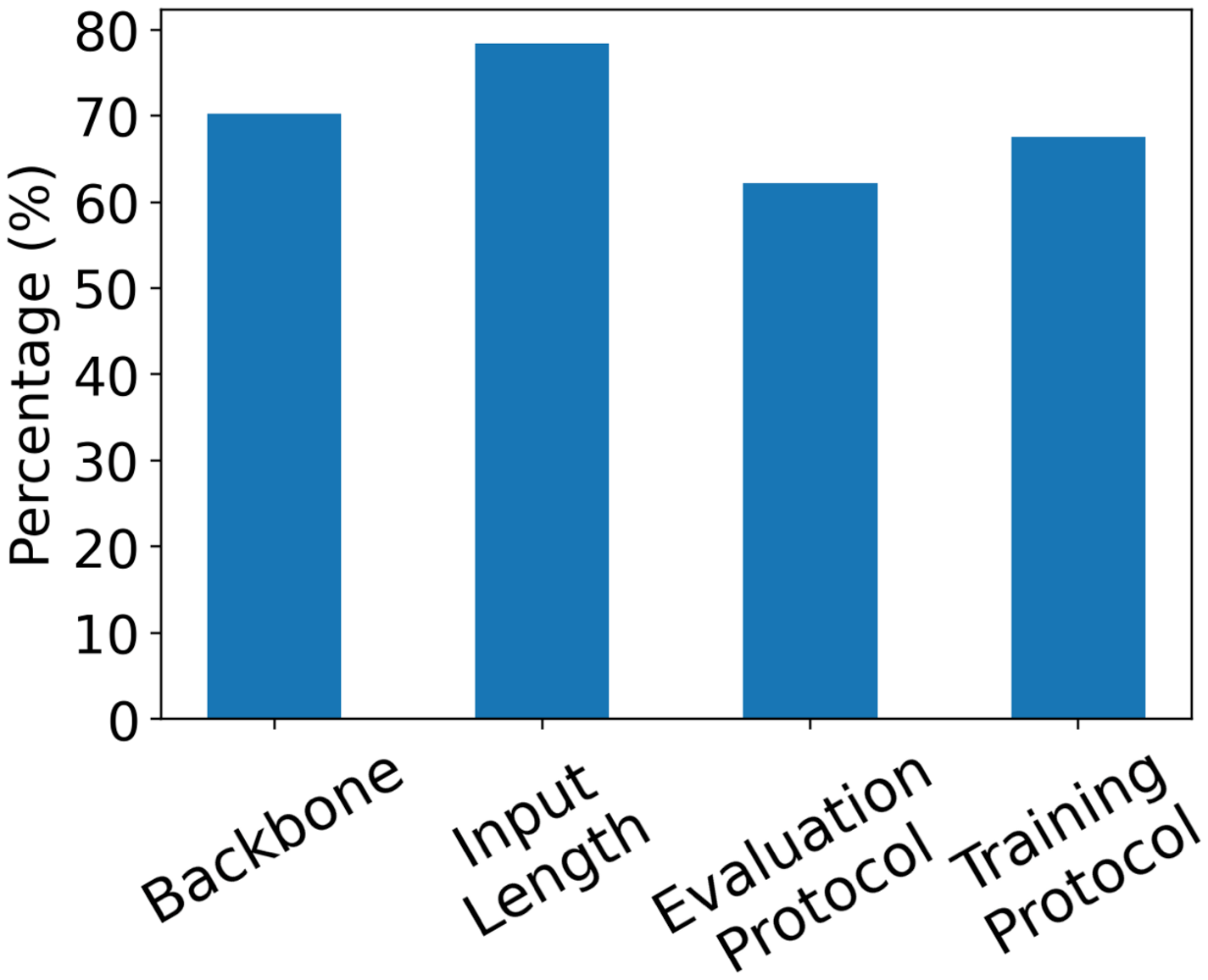}
    \end{subfigure}
    \vspace{-3mm}
    \caption{\small Statistics collected from 37 action recognition papers from 2015 to 2020. Left: Used datasets. Right: Ratio of papers used different settings to compare with others.}
    \label{fig:challenge} \vspace{-5mm}
\end{figure}

The first challenge in evaluating action models stem from the fact that unlike ImageNet for image classification, action recognition does not have one dataset widely used for every paper.
As shown in Figure~\ref{fig:challenge}, the most popular Kinetics-400 is used by around 60\% papers\footnote{Kinetics-400 dataset is available after 2017, the used rate increases to 69\% if only the papers published after 2017 are counted.}. 
On the other hand, Something-Something (V1 and V2), which has very different temporal characteristic from Kinetics-400, is also used by about 50\% papers. Furthermore, two successors of the Kinetics-400 datasets, Kinetics-600 and Kinetics-700 are released recently. It is difficult to evaluate different methods if they do not test on common datasets.
We further check those 37 papers how do they compare the performance in their paper~\cite{C3D:Tran2015learning,TSN:wang2016temporal,I3D:carreira2017quo,ActionVLAD:Girdhar_2017_CVPR,P3D:Qiu_2017_ICCV,MiCT:Zhou_2018_CVPR,OFGuide:Sun_2018_CVPR,TRN:zhou2018temporal,STC:Diba2018spatio,Trajectory:10.5555/3327144.3327148,ResNet3D_2:hara2018can,ARTNet:Wang2018appearance,MFN:10.1007/978-3-030-01249-6_24,ECO:zolfaghari2018eco,S3D,R2plus1D:Tran_2018_CVPR,Nonlocal:Wang2018NonLocal,StNet:He:2019aaai,STD:Martinez_2019_ICCV,STM:Jiang_2019_ICCV,RepFlow:Piergiovanni_2019_CVPR,TSM:lin2018temporal,bLVNetTAM,CollaborativeST:Li_2019_CVPR,Timeception:Hussein_CVPR_2019,GST:Luo_2019_ICCV,SlowFast:feichtenhofer2018slowfast,CSN:Tran_2019_ICCV,TDRL:Weng_2020_ECCV,TEINet:Liu_Luo_Wang_Wang_Tai_Wang_Li_Huang_Lu_2020,sudhakaran2020gate,AssembleNet:Ryoo2020AssembleNet_ICLR2020,TPN:Yang_2020_CVPR,RubiksNet:fanbuch2020rubiks,CorrNet:Wang_2020_CVPR,X3D_Feichtenhofer_2020_CVPR,CIDC:Li_2020_ECCV}. We evaluate those papers from four aspects, including backbone, input length, training protocol and evaluation protocol. Figure \ref{fig:challenge} shows the summary of how papers compare to others differently.

\vspace{1mm}
\noindent \textbf{Backbone.} From our analysis, we observe that about 70\% papers compare results with different backbones (e.g., most of the papers use ResNet50 as backbone but compare with I3D~\cite{I3D:carreira2017quo} which uses InceptionV1 as the backbone). 
Comparing action models with different types of backbones can often lead to incorrect conclusions, also making harder to evaluate the advantage of the proposed temporal modeling. For example, using stronger backbone for I3D, it improves the results by 4.0\% on Kinetics-400 (see Figure~\ref{fig:i3ds_k400}).

\vspace{1mm}
\noindent \textbf{Input Length.} Figure \ref{fig:challenge} shows that about 80\% of the papers use different number of frames for comparison. It is because each method could prefer different frame numbers; however, comparing under different number of frames could favor either the proposed method or the reference method.

\vspace{1mm}
\noindent \textbf{Training Protocol.} Due to recent advances in technology, It is often easier to train action recognition models for a very long time (epochs) which was not possible a few years ago, indicating that old methods might not be well-trained. Furthermore, many works reuse the ImageNet weights to initialize the models while others are not. It raises the concern that does the gain comes from different training protocol. Based on our analysis, about 60\% of the papers use different protocols to train action recognition models. 


\vspace{1mm}
\noindent \textbf{Evaluation Protocol.} As the models are trained under different sampling strategies and input lengths, a model is used to take more than one clip from a video for prediction. Hence, different evaluation protocol could lead unclear comparison. About 60\% papers evaluated models differently when comparing to others.

\section{2D-CNN and 3D-CNN Approaches} 
\label{sec:spatiotemporal}

To address the above mentioned issue for fair comparison, we analyze several popular 2D-CNN and 3D-CNN approaches for action recognition, including I3D~\cite{I3D:carreira2017quo}, ResNet3D~\cite{ResNet3D:hara2017learning}, S3D~\cite{S3D}, R(2+1)D~\cite{R2plus1D:Tran_2018_CVPR}, TSN~\cite{TSN:wang2016temporal} and TAM~\cite{bLVNetTAM}. These approaches not only yield competitive results on popular large-scale datasets, but also widely serve as fundamental building blocks for many other successive approaches such as SlowFast~\cite{SlowFast:feichtenhofer2018slowfast} and CSN~\cite{CSN:Tran_2019_ICCV}. 

Among these approaches, I3D and ResNet3D are pure 3D-CNN models, differing only in backbones. S3D and R(2+1)D factorize a 3D convolutional filter into a 2D spatial filter followed by a 1D temporal filter. In such a sense, they are architecturally similar to 2D models. However, we categorize them into 3D-CNN models as their implementations are based on 3D convolutions. 
While TSN rely only on 2D convolution without temporal modeling, 
TAM, another 2D-CNN approach, adds efficient depthwise temporal aggregation on top of TSN, which shows strong results on Something-Something~\cite{bLVNetTAM}. Finally, since SlowFast is arguably one of the best approaches on Kinetics, we use it as a reference to SOTA results.
Apart from using different types of convolutional kernels, 2D and 3D models differ in a number of other aspects, including model input, temporal pooling, and temporal aggregation, as shown in Table~\ref{table:overview-network}. 

\begin{table}[tb]
    \centering
    \begin{adjustbox}{max width=\linewidth}
    \begin{tabular}{c|c|c|c|c|c|c|c}
        \toprule
            
            \multirow{2}{*}{Approach} & Model & Input &  \multirow{2}{*}{Backbone} & Temporal & Spatial & Temporal & Initial  \\
                                      & Input & Sampling &                           & Pooling  & Module  & Aggregation  & Weights \\
            \midrule
            I3D~\cite{I3D:carreira2017quo} & \multirow{4}{*}{4D} & \multirow{4}{*}{Dense} & InceptionV1 & \multirow{4}{*}{Y} & \multirow{2}{*}{3D Conv.} & \multirow{2}{*}{3D Conv.} & \multirow{2}{*}{Inflation} \\
            
            R3D~\cite{ResNet3D:hara2017learning} & & & ResNet & & & & \\
            \cmidrule{1-1}\cmidrule{4-4}\cmidrule{6-8}
            S3D~\cite{S3D} & & &  InceptionV1 & & \multirow{2}{*}{2D Conv.} & \multirow{2}{*}{1D Conv.} & Inflation \\
            R(2+1)D~\cite{R2plus1D:Tran_2018_CVPR} & & & ResNet & &  &  & Scratch \\
            \midrule
            TAM~\cite{bLVNetTAM} & \multirow{2}{*}{3D} & \multirow{2}{*}{Uniform} & bLResNet & \multirow{2}{*}{N} & \multirow{2}{*}{2D Conv.} & 1D dw Conv. & ImageNet \\
            TSN~\cite{TSN:wang2016temporal} &  & & InceptionV1 & & & None & ImageNet \\
        \bottomrule
    \end{tabular}
    \end{adjustbox}
    \vspace{-2mm}
    \caption{\small 2D-CNN and 3D-CNN approaches in our study.}
    \label{table:overview-network} \vspace{-4mm}
\end{table}

\begin{figure}[t]
    \centering
    \includegraphics[width=.9\linewidth]{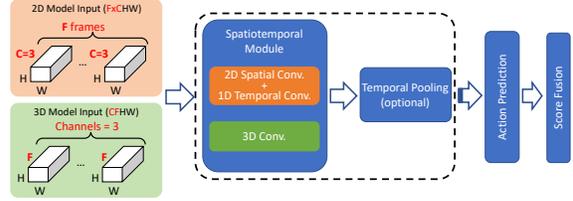} \vspace{-2mm}
    \caption{\small A general framework for 2D-CNN and 3D-CNN approaches of video action recognition. A video action recognition model can be viewed as a sequence of stacked spatio-temporal modules. The input frames are formed as a 3D tensor for 2D models and 4D tensor for 3D models.}
    \label{fig:2d-3d-arch} \vspace{-5mm}
\end{figure}

The differences between 2D-CNN and 3D-CNN approaches make it a challenge to compare these approaches. To remove the bells and whistles and ensure a fair comparison, we show in Figure~\ref{fig:2d-3d-arch} that 2D and 3D models can be represented by a general framework. Under such a framework, an action recognition model is viewed as a sequence of stacked spatio-temporal modules with temporal pooling optionally applied. Thus what differentiates a model from another boils down to only its spatio-temporal module. We re-implemented all the approaches used in our comparison under this framework, which allows us to test an approach flexibly using different configurations such as backbone, temporal pooling and temporal aggregation. For example, in S3D-ResNet (i.e., R(2+1)D), we do not expand the channel dimension between spatial and temporal convolution to keep it align to S3D~\cite{S3D}. More details on the models and implementations can be found in the Supplemental.

\section{Datasets, Training, Evaluation Protocols} 
\label{sec:protocol}

To ensure fair comparison and facilitate reproduciblity, we train all the models using the same data preprocessing, training protocol, and evaluation protocol.
Below we provide a brief description and refer the reader to the Supplemental for more details including the source codes.

\vspace{1mm}
\noindent
\textbf{Datasets.}
We choose Something-Something V2 (\SSV2), Kinetics-400 (\Kinetics) and Moments-in-time (\Moments) for our experiments. We also create a mini version of each dataset: \mSSV2 and \mKinetics account for half of their full datasets by randomly selecting half of the categories of \SSV2 and \Kinetics. \mMoments is provided on the official \Moments website, consisting of 1/8 of videos in the full dataset. 

\vspace{1mm}
\noindent
\textbf{Training.} 
Following~\cite{bLVNetTAM}, we progressively train the models using different input frames. Let $K_i \in [8, 16, 32, 64]$ where $i=1 \hdots 4$. We first train a starter model using 8 frames. The model is either inflated with (e.g., I3D) or initialized from (e.g., TAM) its corresponding ImageNet pre-trained model. We then finetune the model using more frames $K_i$ from the model using $K_{i-1}$ frames. 

\vspace{1mm}
\noindent
\textbf{Evaluation.}
There are two major evaluation metrics for video action recognition: clip-level accuracy and video-level accuracy. Clip-level accuracy is prediction by feeding a single clip into the network and video-level accuracy is the combined predictions of \textit{multiple} clips; thus, the video-level accuracy is usually higher than the clip-level accuracy. By default, we report the clip-level accuracy.

\section{Experimental Results and Analysis}
\label{sec:expr}

In this section, we provide a detailed analysis on the performance of 2D and 3D models (Section~\ref{subsec:2d-3d-performance}), SOTA results and transferability (Section~\ref{subsec:SOTA}) and their spatio-temporal effects (Section~\ref{subsec:spatiotemporal_effects}). 
For clarity, from now onwards, we refer to each of \textit{I3D}, \textit{S3D} and \textit{TAM} as one type of spatio-temporal module illustrated in Figure~\ref{fig:2d-3d-arch}. We name a specific model by \textit{module-backbone[-tp]} where \textit{tp} indicates that temporal pooling is applied. For example, I3D-ResNet18-tp is a 3D model based on ResNet18 with temporal pooling.
To verify the correctness of our implementation, we trained a I3D-InceptionV1 as the original paper~\cite{I3D:carreira2017quo}, and find that our model achieves 73.1\% top-1 accuracy, which is 2\% better than the result reported in the original paper. It clearly justifies the results conducted by our setup is reliable. 

\subsection{Performance Analysis on Mini Datasets}
\label{subsec:2d-3d-performance}
\begin{figure}[t]
    \vspace{-10pt}
    \centering
    \includegraphics[width=\linewidth]{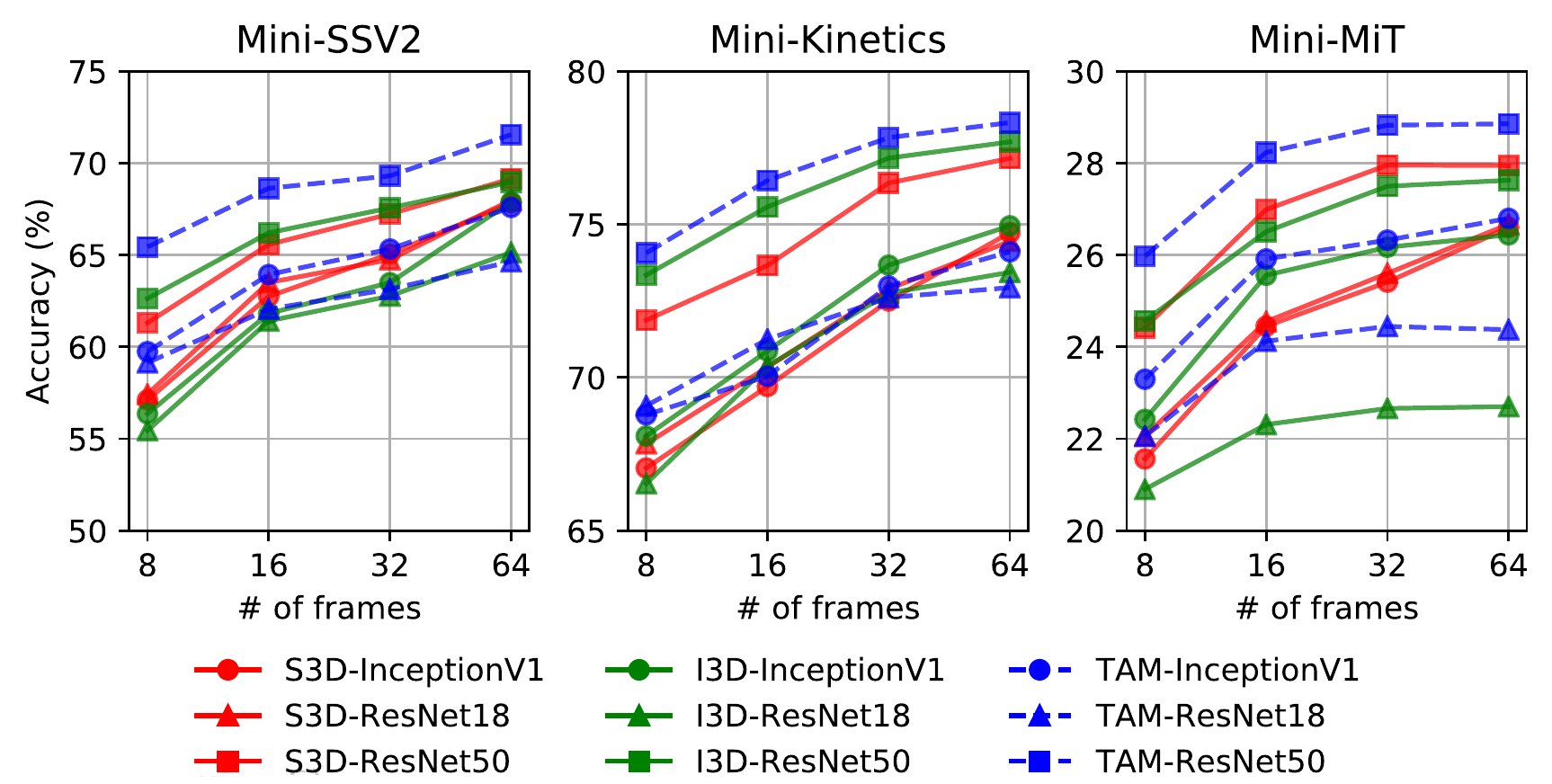}
    \caption{\small Top-1 accuracy of all the compared models without temporal pooling on three mini-datasets. The video architectures are separated by color while the backbones by symbol.}
    \label{fig:all_results} \vspace{-3mm}
\end{figure}

For each spatio-temporal module, we experiment with 3 backbones (InceptionV1, ResNet18 and ResNet50) and two scenarios (w/ and w/o temporal pooling) on three datasets. In each case, 8, 16, 32 and 64 frames are considered as input. This results in a total of $4\times3\times2\times3\times4=288$ models to train, many of which haven't been explored in the original papers. Since temporal pooling is detrimental to model performance (see Figure~\ref{fig:temporal_pooling}), our analysis in this work mainly focus on models w/o temporal pooling unless otherwise specified. Figure~\ref{fig:all_results} reports the clip-level top-1 accuracies w/o temporal pooling for all models. We refer readers to the Supplemental for the results w/ temporal pooling.

\begin{figure}[tb!]
    \centering
    \includegraphics[width=\linewidth]{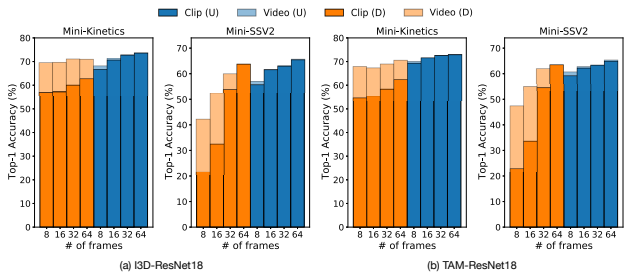} \vspace{-6mm}
    \caption{\small Performance comparison between \textit{Uniform Sampling (U)} and \textit{Dense Sampling (D)}.
    (a) I3D-ResNet18 (b) TAM-ResNet18. Both models do not include temporal pooling. Solid bars are the clip-level accuracy while transparent bars indicates the improvement by the video-level (multi-clip) evaluation.}
    \label{fig:UvsD} \vspace{-5mm}
\end{figure}

\vspace{1mm}
\noindent\textbf{Backbone Network and Input Length.}
As seen from Figure~\ref{fig:all_results}, regardless of the spatiotemporal modules used, there is a general tendency that ResNet50 $>$ InceptionV1 $>$ ResNet18 w.r.t their overall spatiotemporal representation capability. 
Longer input frames tend to produce better results; however, the performance improvement does not seem significant after 32 frames on all three datasets. 

\vspace{1mm}
\noindent \textbf{Input Sampling.} 
Two sampling strategies are widely adopted in action recognition to create model inputs. The first one, \textit{Uniform sampling}, which is often seen in 2D models, divides a video into multiple equal-length segments and then randomly selects one frame from each segment. The other method used by 3D  models, \textit{dense sampling}, instead directly takes a set of continuous frames as the input.
It is not clear, though, why these two types of models prefer different inputs.
Figure~\ref{fig:UvsD} shows that uniform sampling (blue) yields better clip-level accuracies than dense sampling (orange) under all circumstances. This is not surprising as dense sampling only uses part of the test video in the clip-level evaluation. Even though the video-level evaluation boosts the performance of dense sampling by 6\%$\sim$15\% on \mKinetics and 5\%$\sim$20\% on \mSSV2, its computational needs are increased proportionally, e.g., 10 clips used in Figure~\ref{fig:UvsD} to get video-level accuracy, increases the FLOPs by ten folds. Such costs make it inappropriate in practice. Thus, all our analysis is based on uniform sampling and clip-level evaluation unless otherwise stated. We will further analyze the effect of input sampling strategies in Section~\ref{subsec:SOTA} based on the results from full datasets.

\vspace{1mm}
\noindent \textbf{Temporal Pooling.}
Temporal pooling is usually applied to 3D models to reduce computational complexity. It is known that temporal pooling negatively affects model performance. Such effects, however, have not been well understood in the literature. Figure~\ref{fig:temporal_pooling} shows the performance gaps between models w/ and w/o temporal pooling across different backbones and architectures. As can be seen, temporal pooling in general counters the effectiveness of temporal modeling and hurts the performance of action models, just like what spatial pooling does to object recognition and detection. For this reason, more recent 3D-CNN approaches such as SlowFast~\cite{SlowFast:feichtenhofer2018slowfast} and X3D~\cite{X3D_Feichtenhofer_2020_CVPR} drop temporal pooing and rely on other techniques for reducing computation. Similarly, one important reason for the prior finding in~\cite{hutchinson2020accuracy} that 3D models are inferior to C2D (pure spatial models) on \Kinetics and \Moments is because their comparisons neglect the negative impact of temporal pooling on 3D models. As shown in Section~\ref{subsec:SOTA}, I3D w/o temporal pooling is competitively comparable with the SOTA approaches.

\begin{figure}[t!]
    \centering
    \includegraphics[width=1.0\linewidth]{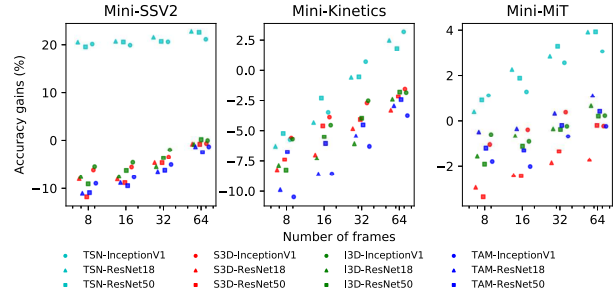} \vspace{-3mm}
    \caption{\small Accuracy gain of the models with temporal pooling w.r.t. the models without temporal pooling. Temporal pooling significantly hurts the performance of all models except TSNs.}
    \label{fig:temporal_pooling} \vspace{-5mm}
\end{figure}

Interestingly, TSN is the only architecture benefiting from temporal pooling, demonstrating a large boost in performance on \mSSV2 ($>$20\%) and \mMoments (3\%$\sim$5\%). Also, as the number of input frames increases, the improvement is more pronounced. On \mKinetics, even though TSN is also negatively affected by temporal pooling , it suffers the least and starts seeing positive gains after 32 frames. To further confirm that, we trained a 32-frame TSN model with temporal pooling on Kinetics. This model (TSN-R50$*$ in Figure~\ref{fig:perf-comparison}) achieves a top-1 accuracy of 74.9\%, 5.1\% higher than the version w/o temporal pooling and only about 2.0\% shy from the SOTA results. We interpret \textit{temporal pooling as a simple form of exchanging information across frames}, which empowers TSN with the ability of temporal modeling. 
The consistent improvements by temporal pooling across all the datasets provide strong evidence that temporal modeling is necessary for video action recognition, even for datasets like Kinetics where temporal information has been shown less crucial for recognition. 




\subsection{Benchmarking of SOTA Approaches}
\label{subsec:SOTA}

\noindent \textbf{Results on Full Datasets.}
I3D based on InceptionV1 has been used as an important baseline by many papers to showcase their progress. However, the results of I3D on the mini datasets, especially the unexpectedly significant impact of temporal pooling, seem to suggest that the spatio-temporal modeling capability of I3D has been underestimated by the field. To more precisely understand the recent progress in action recognition, we further conduct a more rigorous benchmarking effort including I3D, TAM and SlowFast on the full datasets. I3D was the prior SOTA method while SlowFast~\cite{SlowFast:feichtenhofer2018slowfast} and TAM~\cite{bLVNetTAM}, both of which have official codes released, are competitively comparable with existing SOTA methods. 
 To ensure apple-to-apple comparison, we follow the same training settings of SlowFast to train all the models using 32 frames as input. 
During evaluation, we use $3\times10$ clips for \Kinetics and \Moments, and $3\times2$ clips for \SSV2.

\begin{figure}[t!]
    \centering
    \includegraphics[width=.8\linewidth]{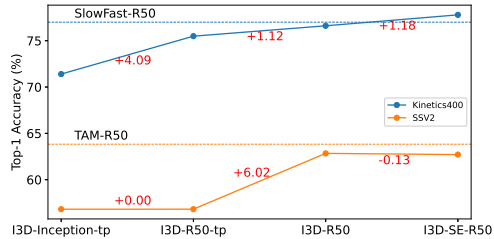} \vspace{-2mm}
    \caption{\small Performance of I3D models by changing the backbone (I3D-R50-tp), removing temporal pooling (I3D-R50-tp) and adding squeeze-excitation modules (I3D-SE-R50) on \Kinetics and \SSV2. Red numbers indicate performance changes. All models are trained with 32 frames and evaluated using 3 $\times$ 10 clips on \Kinetics, and 3 $\times$ 2 clips on \SSV2, respectively.}
    \label{fig:i3ds_k400} \vspace{-4mm}
\end{figure}

We first augment original I3D by a stronger backbone ResNet50 and turning off temporal pooling. As shown in Figure~\ref{fig:i3ds_k400}, ResNet50 alone pushes up the accuracy of I3D by 4.0\% on \Kinetics, and removing temporal pooling adds another 1.1\% performance gain, putting I3D on par with SlowFast in terms of top-1 accuracy. Further inserting Squeeze-Excitation modules into I3D makes it surpass SlowFast by 0.8\%. On \SSV2, a stronger backbone provides I3D little benefit in accuracy, but removing temporal pooling boosts the performance substantially by 6\%, making I3D comparable to TAM. 
Table~\ref{table:sota} provides more detailed results in this experiment. In summary, I3D-ResNet50 demonstrates impressive results, staying on par with state-of-the-art approaches in accuracy on all three datasets. The fact that I3D remains very strong across multiple large-scale datasets suggests that the recent progress of action recognition in terms of accuracy is largely attributed to the use of more powerful backbone networks, but not the improved spatio-temporal modeling as expected.  Nevertheless, we do observe that recent approaches such as X3D~\cite{X3D_Feichtenhofer_2020_CVPR} have made a large leap ahead in efficiency (FLOPs) compared to I3D.
Moreover, SlowFast performs worse than I3D and TAM on \SSV2 on the Something-Something dataset. We speculate that this could be related to: (I) that the slow pathway only uses temporal convolutions after stage4 of ResNet, which weakens its temporal modeling capability; and (II) that the two-stream architecture is less effective in capturing temporal dependencies in such a highly temporal dataset.

\begin{table}[tb]
    \centering
    \begin{adjustbox}{max width=\linewidth}
    \begin{tabular}{c|c|c|c|c|c}
        \toprule
            \multirow{2}{*}{Model} & Pretrain & \multirow{2}{*}{FLOPs} & \multicolumn{3}{c}{Dataset}\\
                                      & dataset & &  \Kinetics & \SSV2 & \Moments \\
            \midrule
            I3D-ResNet50 & ImageNet & 335.3G & 76.61  & 62.84 & 31.21 \\ 
            I3D-ResNet50 & None & 335.3G & 76.54  & $-$ & $-$  \\ 
            TAM-ResNet50 & ImageNet & 171.5G & 76.18 & 63.83 & 30.80  \\ 
            SlowFast-ResNet50-8$\times$8$^\dagger$~\cite{SlowFast:feichtenhofer2018slowfast} & None$^*$ & 65.7G & 76.40 & 60.10 & 31.20 \\
            \midrule
            I3D-ResNet101 & ImageNet & 654.7G & 77.80 & 64.29 & $-$ \\
            TAM-ResNet101 & ImageNet & 327.1G & 77.61 & 65.32 & $-$ \\
            \midrule
            SlowFast-ResNet50-8$\times$8$^{\ddagger}$~\cite{SlowFast:feichtenhofer2018slowfast} & None$^*$ & 65.7G & 77.00 & $-$ & $-$ \\
            SlowFast-ResNet50-16$\times$8$^{\ddagger}$~\cite{SlowFast:feichtenhofer2018slowfast} & \Kinetics & 124.5G & $-$ & 63.0 & $-$ \\
            CorrNet-ResNet50$^{\ddagger}$~\cite{CorrNet:Wang_2020_CVPR} & None$^*$ & 115G & 77.20 & $-$  & $-$ \\
            SlowFast-ResNet101-8$\times$8$^{\dagger}$~\cite{SlowFast:feichtenhofer2018slowfast} & None & 125.9G & 76.72 & $-$ & $-$ \\
            SlowFast-ResNet101-8$\times$8$^{\ddagger}$~\cite{SlowFast:feichtenhofer2018slowfast} & None & 125.9G & 78.00 & $-$ & $-$ \\
            SlowFast-ResNet101-16$\times$8$^{\ddagger}$~\cite{SlowFast:feichtenhofer2018slowfast} & None & 213G & 78.90 & $-$ & $-$ \\
            CSN-ResNet101$^{\ddagger}$~\cite{CSN:Tran_2019_ICCV} & None$^*$ & 83G & 76.70 &  $-$ & $-$  \\
            CorrNet-ResNet101$^{\ddagger}$~\cite{CorrNet:Wang_2020_CVPR} & None$^*$ & 224G & 79.20 & $-$ & $-$  \\
            X3D-L$^{\ddagger}$~\cite{X3D_Feichtenhofer_2020_CVPR} & None$^*$ & 24.8G & 77.50 & $-$ & $-$  \\
            X3D-XL$^{\ddagger}$~\cite{X3D_Feichtenhofer_2020_CVPR} & None$^*$ & 48.4G & 79.10 & $-$ & $-$  \\
            \midrule
            \midrule
            AssembleNet-50$^1$~\cite{AssembleNet:Ryoo2020AssembleNet_ICLR2020} & $-$ & $-$ & $-$ & $-$ & 31.41 \\
            GST-ResNet101~\cite{CollaborativeST:Li_2019_CVPR} & ImageNet & $-$ & $-$ & $-$ & 32.40 \\
        \bottomrule
        \multicolumn{6}{l}{\footnotesize{$^*$: Those networks cannot be initialized from ImageNet due to its structure.}} \\
        \multicolumn{6}{l}{\footnotesize{$^\dagger$: Retrained by ourselves. $^{\ddagger}$: reported by the authors of the paper. $^1$: Use RGB + Flow.}}
    \end{tabular}  
    \end{adjustbox}
    \vspace{-3mm}
    \caption{\small Performance of SOTA models.}
    \vspace{-1mm}
    \label{table:sota}
\end{table}

\begin{table}[tb]
    \centering
    \begin{adjustbox}{max width=.9\linewidth}
    \begin{tabular}{c|c|cc}
        \toprule
            Model & Pretrain & U-Sampling & D-Sampling \\
            \midrule
            I3D-ResNet50 & ImageNet & 76.07 &76.61\\ 
            TAM-ResNet50 & ImageNet & 76.45 & 76.18\\ 
            SlowFast-ResNet50-8$\times$8 & $-$ & 71.85 &76.40 \\
        \bottomrule
    \end{tabular}  
    \end{adjustbox}
    \vspace{-2mm}
    \caption{\small Model performance on \Kinetics based on uniform and dense sampling. Uniform sampling trained models are evaluated under 3 256$\times$256 spatial crops and 2 clips.}
    \label{table:sota_uniform} \vspace{-4mm}
\end{table}

\vspace{1mm}
\noindent\textbf{Uniform Sampling vs Dense Sampling.}
We revisit the effect of input sampling on model performance and retrain all three approaches using uniform sampling on~\Kinetics. As shown in Table~\ref{table:sota_uniform}, the small difference between uniform and dense sampling results indicates that both I3D and TAM are flexible w.r.t model input. In contrast, uniform sampling is not as friendly as dense sampling to SlowFast, producing an accuracy $\sim$5\% lower than dense sampling. We conjecture that this has to do with dual-path architecture of SlowFast. Such an architecture is primarily designed for efficiency and possibly less effective in learning spatial-temporal representations from sparsely sampled frames (i.e. 8-frame uniform sampling in this case). This also explains why SlowFast, when trained with uniform sampling, under performs by 2\% $\sim$ 3\% on \SSV2 in Table~\ref{table:sota} in contrast to I3D and TAM.

\begin{figure}[t!]
    \centering
    \begin{subfigure}{0.49\linewidth}
      \includegraphics[width=1\linewidth]{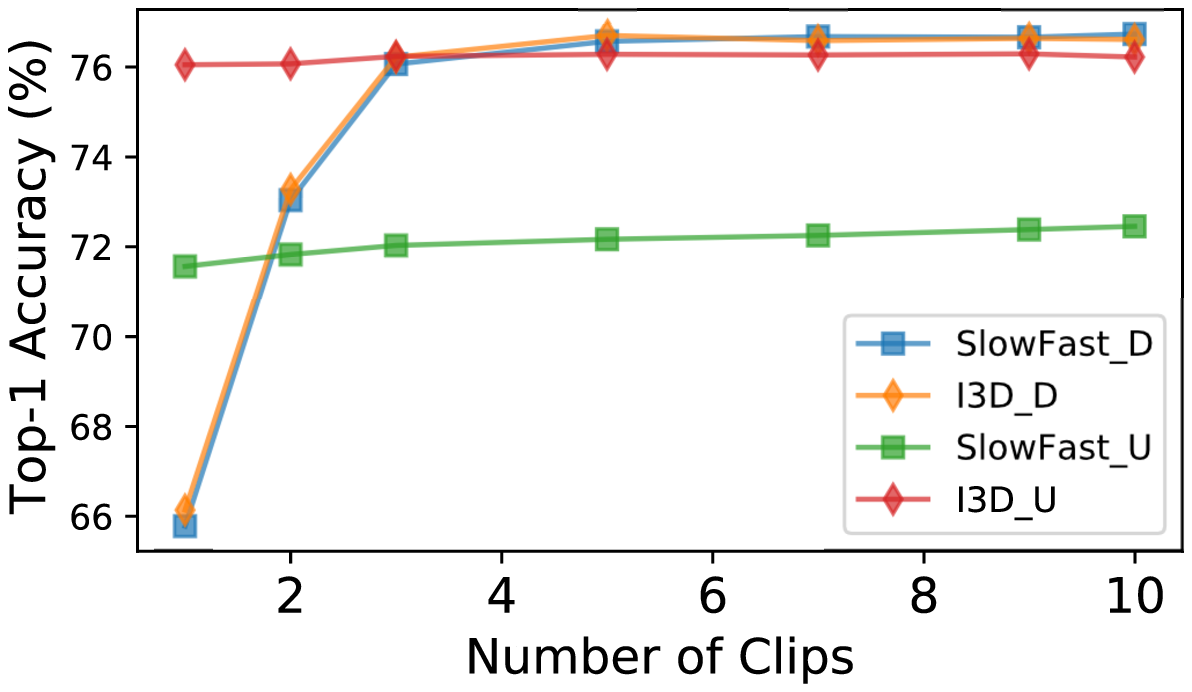}
      \vspace{-4mm}
    \label{fig:diff_num_clips-a}
    \end{subfigure}
    \hfill
    \begin{subfigure}{0.49\linewidth}
      \includegraphics[width=1\linewidth]{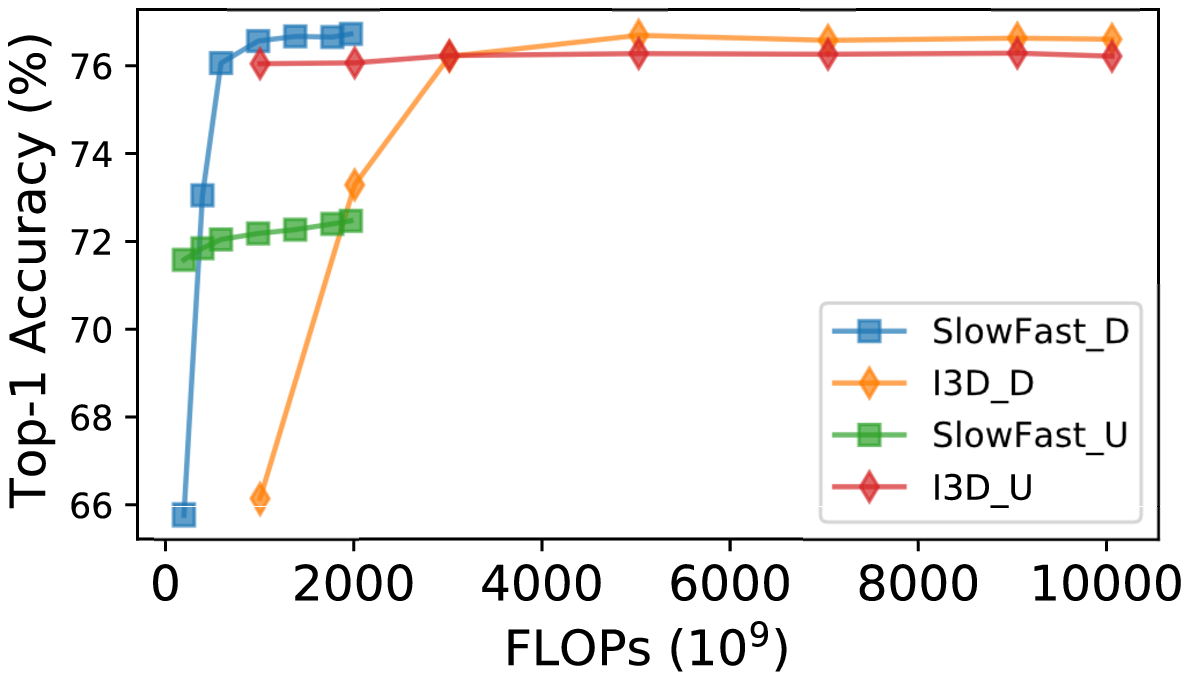}
      \vspace{-4mm}
    \label{fig:diff_num_clips_flops-b}
    \end{subfigure}
    \vspace{-3mm}
    \caption{\small Model performance tested using 3 256$\times$256 spatial crops and different number of clips. 'U': uniform sampling; 'D': dense sampling. Best viewed in color. 
    }
    \label{fig:diff_num_clips} 
\end{figure}

Furthermore, Figure~\ref{fig:diff_num_clips} (Left) shows model accuracy v.s. number of clips used for evaluation in uniform and dense sampling, respectively. As can be observed, the model performance with dense sampling is saturated quickly after 4-5 clips for both I3D and SlowFast. This suggests that the common practice in the literature of using $10$ clips for dense sampling is often not necessary. As opposed to dense sampling, uniform sampling benefits slightly (i.e., for SlowFast) or little from multiple clips. This raises another pitfall that is largely overlooked by the community when assessing model efficiency, i.e., the impact of input sampling. As shown in Figure~\ref{fig:diff_num_clips} (Right), when putting I3D and SlowFast in a plot of accuracy v.s. FLOPs for comparison, the advantage of SlowFast over I3D is better and more fairly represented, i.e., when considering uniform sampling for I3D, SlowFast is only slightly more accurate but at the same efficiency in FLOPs. This clearly suggests that input sampling strategy of a model (i.e. uniform or dense) should factor in evaluation for fairness when comparing it to another model.

\vspace{1mm}
\noindent \textbf{Model Transferability.}
We further compare the transferability of the three models trained above on four small-scale datasets including UCF101~\cite{ucf101:Soomro2012}, HMDB51~\cite{HMDB:Kuehne2011}, Jester~\cite{Materzynska_2019_ICCV}, and \mSSV2.
We follow the same training setting in Section~\ref{sec:protocol} and finetune 45 epochs with cosine annealing learning rate schedule starting with 0.01; furthermore, since those are 32-frame models, we trained the models with a batch size of 48 with synchronized batch normalization. 
Table~\ref{table:transfer} shows the results, indicating that all the three models have very similar performance (difference of less than 2\%) on the downstream tasks. In particular, I3D performs on par with the SOTA approaches like TAM and SlowFast in transfer learning (e.g., I3D obtains the best accuracy of 97.12\% on UCF101), which once again corroborates the fact that the improved spatio-temporal modeling is largely due the use of stronger backbones. 
\label{subsec:transferability}
\begin{table}[tb]
    \centering
    \begin{adjustbox}{max width=.9\linewidth}
    \begin{tabular}{c|c|c|c|c}
        \toprule
              & \multicolumn{4}{c}{Target dataset}\\
               Model                   & UCF101 & HMDB51 & Jester & \mSSV2 \\
            \midrule
             I3D-ResNet50 & 97.12  & 72.32 & 96.39 & 65.86  \\
             TAM-ResNet50 &  95.05& 71.67 & 96.35 & 66.91 \\
             SlowFast-ResNet50-8$\times$8 &  95.67 & 74.61 & 96.75 & 63.93  \\
        \bottomrule
    \end{tabular}  
    \end{adjustbox}
    \vspace{-2mm}
    \caption{\small Top-1 Acc. of Transferability study from \Kinetics.}
    \label{table:transfer} \vspace{-3mm}
\end{table}


\subsection{Analysis of Spatio-temporal Effects}
\label{subsec:spatiotemporal_effects}

It's generally believed that temporal modeling is the core for action recognition and state-of-the-art approaches can capture better temporal information. However, it has also been demonstrated on datasets such as \Kinetics and \textit{Moments-in-Time} (\Moments)~\cite{Moments:monfort2019moments} that approaches purely based on spatial modeling~\cite{TSN:wang2016temporal, Moments:monfort2019moments} can achieve very competitive results compared to more sophisticated spatio-temporal models. More recently, a paper~\cite{hutchinson2020accuracy} also shows that 2D  models outperform their 3D counterparts on the \Moments benchmark.
These findings seem to imply that more complex temporal modeling is not necessary for ``static" datasets such as \Kinetics and \Moments.
We believe that lack of fairness in performance evaluation leads to confusion on understanding significance of temporal modeling for action recognition. 

\begin{table}[t]
    \centering
    \begin{adjustbox}{max width=\linewidth}
    \begin{tabular}{c|c|c|c|c|c|c|c|c|c|c|c}
        \toprule
        \multirow{2}{*}{Dataset} & \multirow{2}{*}{Frames} &  \multicolumn{4}{c|}{InceptionV1} & \multicolumn{6}{c}{ResNet50} \\ 
        \cmidrule{3-12}
        & & None & I3D & Conv. &  TAM & None & I3D & Conv. &  TAM &   TSM & NLN \\ 
        \midrule
        \multirow{2}{*}{\mSSV2} & $f$=8  & 33.1 & 56.4 &  58.2 & \textbf{59.7} & 33.9 & 62.6 & 61.6 & \textbf{65.4} & 64.1 & 53.0 \\
         & $f$=16 & 34.7 & 61.8 & 63.7 & \textbf{63.9} & 35.3 & 66.2 & 65.7 & \textbf{68.6} & 67.4 & 55.0 \\
        \midrule
        \multirow{2}{*}{\mKinetics} & $f$=8  & \textbf{70.4} & 68.1 & 68.3 & 68.8 & 72.1 & 73.3 & 71.5 & \textbf{74.1} & \textbf{74.1} & 73.7 \\
         & $f$=16 & 70.5 & 70.9 & \textbf{70.7} & 70.0 & 72.5 & 75.5 & 73.4 & \textbf{76.4} & 75.6 & 74.5 \\
        \bottomrule
    \end{tabular}
    \end{adjustbox}
    \vspace{-1mm}
    \caption{\small Performance of different temporal aggregation strategies w/o temporal pooling. FLOPs and parameters of different models can be found in the supplementary material.}
    \label{table:diff_ta} \vspace{-4mm}
\end{table}

\vspace{1mm}
\noindent
\textbf{Temporal Aggregation.}
\label{subsec:aggregation}
The essence of temporal modeling is how it aggregates temporal information.
The 2D architecture offers great flexibility in temporal modeling. For example, TSM~\cite{TSM:lin2018temporal} and TAM~\cite{bLVNetTAM} can be easily inserted into a CNN for learning spatio-temporal features. Here we analyze several basic temporal aggregations on top of the 2D architecture including 1D convolution (\textit{Conv}, i.e., S3D~\cite{S3D}), 1D depthwise convolution (\textit{dw Conv}, i.e., TAM), and TSM.
We also consider the non-local network module (NLN)~\cite{Nonlocal:Wang2018NonLocal} for its ability to capture long-range temporal video dependencies add 3 NLN modules and 2 NLN modules at stage 2 and stage 3 of TSN-ResNet50, respectively as in~\cite{Nonlocal:Wang2018NonLocal}. 

Table~\ref{table:diff_ta} shows results of using different temporal aggregations as well as those of TSN (i.e., w/o any temporal aggregation) on InceptionV1 and ResNet50. The results suggest that \emph{effective temporal modeling is required for achieving competitive results, even on datasets such as \Kinetics where temporal information is thought as non-essential for recognition.} On the other hand, TAM and TSM, while being simple and efficient, demonstrate better performance than the I3D, 1D regular convolution and the NLN module, which have more parameters and FLOPs. We argue it is because the frames sampled under uniform sampling are sparse and it is not suitable to model temporal information in 3D convolution. While TAM and TSM use depthwise convolution that is more effective to model temporal information since it only consider the single feature map at different frames once instead of combining all channels of frames once. We also find the same pattern on full \Kinetics in Table~\ref{table:sota_uniform}.
Interestingly, the NLN module does not perform as expected on \mSSV2. This is possibly because NLN models temporal dependencies through matching spatial features between frames, which are weak in \mSSV2.



\begin{table}
    \centering
    \begin{adjustbox}{max width=.9\linewidth}
    \begin{tabular}{c|c|cc}
        \toprule
            & & \multicolumn{2}{c}{Top-1 Acc.} \\
        \# of TAMs & locations & \mSSV2 & \mKinetics \\
        \midrule
        8 & All & 59.1 &  69.08\\
        4 & Top-half & 59.7 & 69.21 \\
        4 & Bottom-half & 56.5 & 69.27\\
        4 & Uniform-half & 59.4 & 69.14\\
        \bottomrule
        \multicolumn{4}{l}{\footnotesize{Top and bottom mean the residual blocks closer to output and input respectively.}}\\
    \end{tabular}
    \end{adjustbox}
    \vspace{-2mm}
    \caption{\small Performance comparison by using different numbers and locations of TAMs in ResNet18 (w/o temporal pooling).}
    \label{table:diff_num_ta} \vspace{-4mm}
\end{table}

\vspace{1mm}
\noindent
\textbf{Locations of Temporal Modules.}
\label{subsec:module-location}
In~\cite{S3D} and ~\cite{R2plus1D:Tran_2018_CVPR}, some preliminary analysis w.r.t the effect of the locations of temporal modules on 3D models was performed on Kinetics-400. In this experiment, we conduct a similar experiment on both \mKinetics and \mSSV2 to understand if this is so for 2D models. We modified TAM-ResNet18 in a number of different ways by keeping: a) half of the temporal modules only in the bottom network layers (\textit{Bottom-Half}); b) half of the temporal modules only in the top network layers (\textit{Top-Half}); c) every other temporal module (\textit{Uniform-Half}); and d) all the temporal modules (\textit{All}). As observed in Table~\ref{table:diff_num_ta}, only half of the temporal modules (\textit{Top-Half}) is needed to achieve the best accuracy on \mSSV2 while the accuracy on \mKinetics is not sensitive to the number and locations of temporal modules. It is thus interesting to explore if this insightful observation can lead to an efficient but effective video architecture by mixing 2D and 3D modelings, similar to the idea of ECO in~\cite{ECO:zolfaghari2018eco}.

\vspace{1mm}
\noindent
\textbf{Disentangling Spatial and Temporal Effects.}
\label{subsec:st-disentanglement}
So far we have only looked at the overall spatio-temporal effects of a model (i.e., top-1 accuracy) in our analysis. Here we further disentangle the spatial and temporal contributions of a model to understand its ability of spatio-temporal modeling. Doing so provides great insights into which information, spatial or temporal, is more essential to recognition.
We treat TSN w/o temporal pooling as the baseline spatial model as it does not model temporal information. TSN can evolve into different types of spatio-temporal models by adding temporal modules on top of it. With this, we compute the spatial and temporal contributions of a model as follows.
Let $S_a^b(k)$ be the accuracy of a model of some architecture $a$ that is based on a backbone $b$ and takes $k$ frames as input. For instance, $S_{I3D}^{ResNet50} (16)$ is the accuracy of a 16-frame I3D-ResNet50 model. Then the spatial contribution $\Phi_a^{b}$ and temporal improvement of a model $\Psi_a^{b}$ ($k$ is omitted here for clarity) are given by,

\begin{equation}
\label{eq:sc}    
\begin{aligned}
\Phi_a^{b} = & S_{TSN}^b  / \max{(S_a^b, S_{TSN}^{b})} \\
\Psi_a^{b} = & (S_a^b - S_{TSN}^{b}) / (100 - S_{TSN}^{b}).
\end{aligned}
\end{equation}

Note that $\Phi_a^{b}$ is between 0 and 1;  
$\Psi_a^{b} < 0$ indicates that temporal modeling is harmful to model performance. 
We further combine $\Phi_a^{b}$ and $\Psi_a^{b}$ across all models with different backbone networks to obtain average spatial and temporal contributions of a network architecture, as shown below.
\begin{equation}
 \bar{\Phi}_a  = \frac{1}{Z_{\Phi}}\sum_{b \in B}\sum_{k \in K} \Phi_a^{b}(k), ~~~~~ \bar{\Psi}_a  = \frac{1}{Z_{\Psi}}\sum_{b \in B}\sum_{k \in K} \Psi_a^{b}(k),
\end{equation}
where $B$ = \{InceptionV1, ResNet18, ResNet50\}, $K$ = \{8, 16, 32, 64\}. $Z_{\Phi}$ and $Z_{\Psi}$ are the normalization factors.

\begin{table}[!tb]
     \centering
     \begin{adjustbox}{max width=.8\linewidth}
     \begin{tabular}{c|c|c|c|c}
         \toprule
        
        Datasets & Metrics    & I3D & S3D & TAM      \\
         \midrule
                  &   $\bar{\Phi}_a$      & 0.53 & 0.53 & 0.52   \\
         Mini-SSV2   &   $\bar{\Psi}_a^{ta}$ & 0.46 & 0.45 & 0.47   \\
              &   $\bar{\Psi}_a^{ta+tp}$ & 0.38 & 0.38 & 0.37\\
         \midrule
                  &   $\bar{\Phi}_a$      & 0.97 & 0.97 & 0.96\\
         Mini-Kinetics    &   $\bar{\Psi}_a^{ta}$ & 0.06 & 0.08 & 0.09\\
          &   $\bar{\Psi}_a^{ta+tp}$ &-0.08 &-0.10 &-0.12\\
         \midrule
                  &   $\bar{\Phi}_a$      & 0.89 & 0.91 & 0.87 \\
         Mini-MiT    &   $\bar{\Psi}_a^{ta}$ & 0.04 & 0.03 & 0.04 \\
           &   $\bar{\Psi}_a^{ta+tp}$ & 0.02 & 0.02 & 0.04\\
         \bottomrule
          \multicolumn{5}{l}{\footnotesize{$\bar{\Psi}_a^{ta}$: the improvement from temporal aggregation only.}} \\
          \multicolumn{5}{l}{\footnotesize{$\bar{\Psi}_a^{ta+tp}$: the improvement from combining temporal }} \\
          \multicolumn{5}{l}{\footnotesize{}}
     \end{tabular}
     \end{adjustbox}
     \vspace{-6mm}
     \caption{\small Effects of spatio-temporal modeling.}
     \vspace{-4mm}
     \label{table:spt_modeling} 
\end{table}

Table~\ref{table:spt_modeling} shows the results of $\bar{\Phi}_a $ and $\bar{\Psi}_a $ for three spatio-temporal representations. All three representations behave similarly, namely their spatial modeling contributes slightly more than temporal modeling on \mSSV2, much higher on \mMoments, and dominantly on Mini-Kinetics. This convincingly explains why a model lack of temporal modeling like TSN can perform well on \mKinetics, but fail badly on \mSSV2. Note that similar observations have been made in the literature, but not in a quantitative way like ours. Furthermore, while all the approaches indicate the utmost importance of spatial modeling on mini-Kinetics, the results of $\bar{\Psi}_a^{ta}$ suggest that temporal modeling is more effective on \mKinetics than on \mMoments for both 2D and 3D approaches. 
We also observe that temporal pooling deters the effectiveness of temporal modeling on all the approach from the results of $\bar{\Psi}_a^{ta+tp}$, which are constantly lower than $\bar{\Psi}_a^{ta}$. Such damage is especially substantial on \mKinetics, indicated by the negative values of $\bar{\Psi}_a^{ta+tp}$.

\section{Conclusion}
In this paper, we conducted a comprehensive comparative analysis of several representative CNN-based video action recognition approaches with different backbones and temporal aggregations. Our extensive analysis enables better understanding of the differences and spatio-temporal effects of 2D-CNN and 3D-CNN approaches. It also
provides significant insights with regard to the efficacy of spatio-temporal representations for action recognition.

\vspace{1mm}
{\small 
\noindent
\textbf{Acknowledgments.} This work is supported by the Intelligence Advanced Research Projects Activity (IARPA) via DOI/IBC contract number D17PC00341. The U.S. Government is authorized to reproduce and distribute reprints for Governmental purposes notwithstanding any copyright annotation thereon. This work is also supported by the MIT-IBM Watson AI Lab.

\noindent
\textbf{Disclaimer.} The views and conclusions contained herein are those of the authors and should not be interpreted as necessarily representing the official policies or endorsements, either expressed or implied, of IARPA, DOI/IBC, or the U.S. Government.
}

{\small
\bibliographystyle{ieee_fullname}
\bibliography{reference}
}

\clearpage
\appendix

\noindent \textbf{Summary.}
As part of the supplementary material, we first compare additional SOTA models trained with uniform sampling and discuss runtime speed/memory of different models in Section~\ref{sec:supp:sota}. We analyze the effects of pretrained models in Section~\ref{sec:supp:train_from_scratch}. Moreover, we expand Table~7 of the main paper with the FLOPs and parameters for different temporal aggregations in Table~\ref{table:diff_ta_d}. We then describe the datasets and benchmarks in Section~\ref{sec:supp:datasets} and Section~\ref{sec:supp:training}, respectively. Section~\ref{sec:supp:implementation} describes the model implementation used in the Mini-Datasets benchmark. Figure~\ref{fig:supp:all_results_wtp} shows the results of all the models on Mini datasets. 

\section{More Results on SOTA Models}
\label{sec:supp:sota}

\begin{figure*}[!ht]
    \centering
    \begin{subfigure}{0.49\linewidth}
      \includegraphics[width=1\linewidth]{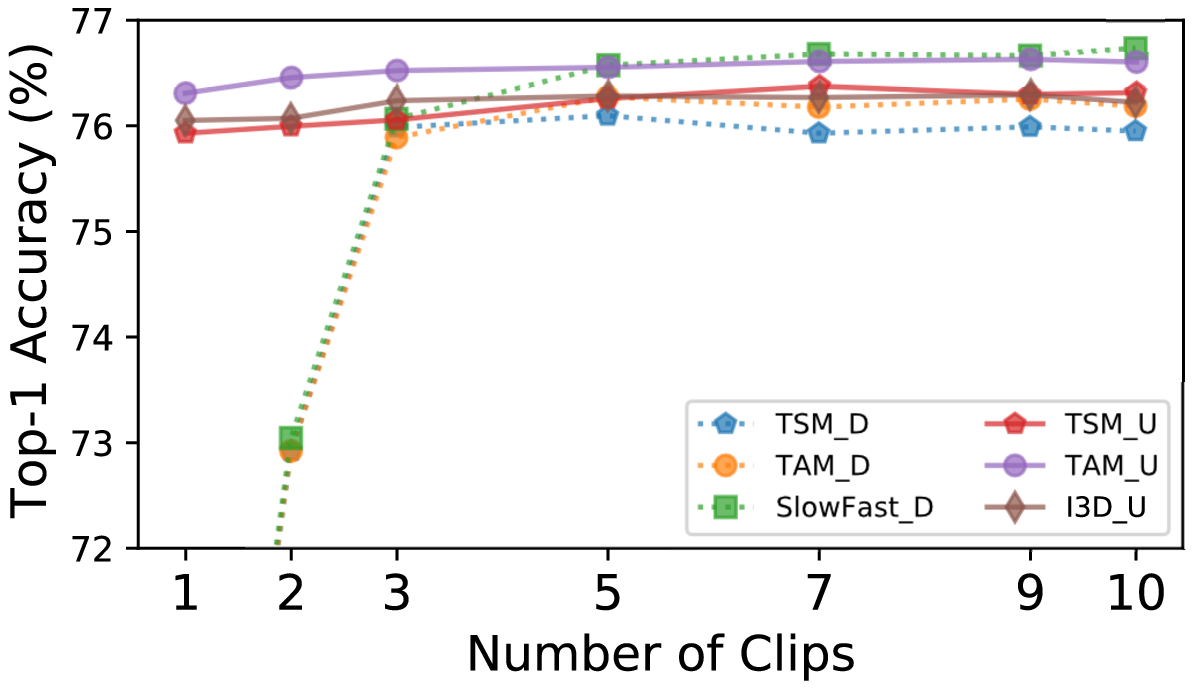}
      \vspace{-4mm}
    \end{subfigure}
    \hfill
    \begin{subfigure}{0.49\linewidth}
      \includegraphics[width=1\linewidth]{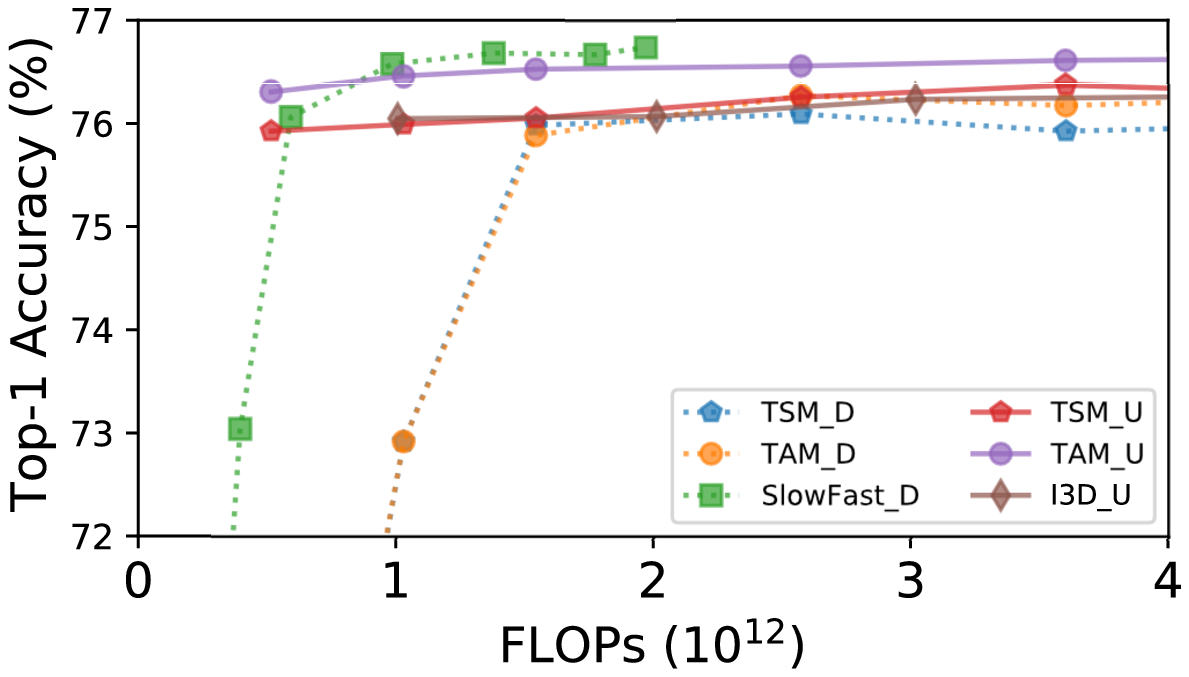}
      \vspace{-4mm}
    \end{subfigure}
    \vspace{-4mm}
    \caption{\small Model performance tested using 3 256$\times$256 spatial crops and different number of clips. 'U': uniform sampling; 'D': dense sampling. Best viewed in color.
    }
    \label{fig:supp:diff_num_clips} \vspace{-2mm}
\end{figure*}

We argue in the main paper that  the effect of input sampling (i.e.,
uniform or dense sampling) should be considered for fair comparison of action recognition models. Following this practice, we add TSM~\cite{TSM:lin2018temporal} and TAM~\cite{bLVNetTAM} to Figure 8 of the main paper, which is shown here as Figure~\ref{fig:supp:diff_num_clips}. 
Like I3D, TAM and TSM work well with both uniform and dense sampling. When uniform sampling is considered for TAM and TSM,  they are slightly worse than SlowFast in accuracy but at the same efficiency in FLOPs.

Moreover, we also benchmark the runtime speed and memory consumption for a more comprehensive comparison among different models.
Table~\ref{table:sota_speed} shows throughput, FLOPs and max batch size that can be fitted on a GPU. The numbers are conducted based on a NVIDIA V100 32G GPU with CUDA 10.1, cudnn 7.6.5, and PyTorch 1.5; and the throughput is measured under maximum batch size.
Slowfast achieved better speed and memory usage; on the other hand, even though TAM-R50 has only half FLOPs of I3D-R50, their throughputs are similar. This is because the 3D-depthwise convolution used in TAM is not yet optimized\footnote{\url{https://github.com/pytorch/pytorch/pull/40801}}.

\begin{table}[tb]
    \centering
    \begin{adjustbox}{max width=\linewidth}
    \begin{tabular}{c|c|c|c|c}
        \toprule
        Models & Accuracy & FLOPs & Throughput  & Maximum  \\
            & (\%) & (G) &  (clips/sec) & batch size \\
        \midrule
        TSN-ResNet50-tp & 74.9 & 73.9 & 53.5 & 40 \\
        TAM-ResNet50 & 76.2 & 171.5 & 14.2 & 48 \\
        I3D-ResNet50 & 76.6 & 335.3 & 16.1 & 44 \\
        SlowFast-ResNet50-8$\times$8 & 76.4 & 65.7 & 48.9 & 96\\
        \bottomrule
    \end{tabular}
    \end{adjustbox}
    \caption{Speed and memory complexity of models.}
    \label{table:sota_speed}
\end{table}

\section{Effects of Pretrained Models}
\label{sec:supp:train_from_scratch}
It is shown in \cite{He_2019_ICCV_rethinking} that when 
there is enough training data and the training is sufficiently long, then pretraining from ImageNet is not necessary for producing competitive performance in a downstream task like  object detection. Here we conduct a similar experiment to train I3D-ResNet50 and TAM-ResNet50 from scratch for 196 epoches. As can be seen from
Table~\ref{table:supp:train_from_scratch}, I3D-ResNet50 does not benefit much from a pretrained ImageNet model. For TAM-ResNet50, when trained from scratch, it degrades its accuracy by $\sim0.5$, which is not significant. 
Thus our results seem to echo a similar observation that pretraining is not crucial for large-scale video action recognition as long as the training is allowed to be sufficiently long.


\begin{table}[t]
    \centering
    \begin{adjustbox}{max width=\linewidth}
    \begin{tabular}{c|c|c}
        \toprule
            \multirow{2}{*}{Model} & \multicolumn{2}{c}{Pretrain} \\
                  & ImageNet & None \\
            \midrule
            I3D-ResNet50 & 76.61 & 76.54 \\
            TAM-ResNet50 & 76.18 & 75.61 \\
        \bottomrule
    \end{tabular}  
    \end{adjustbox}
    \vspace{-1mm}
    \caption{\small Top-1 Accuracy (\%) of I3D and TAM models trained with and without ImageNet weights on \Kinetics.}
    \vspace{-1mm}
    \label{table:supp:train_from_scratch}
\end{table}

\begin{table*}[t]
    \centering
    \begin{adjustbox}{max width=\linewidth}
    \begin{tabular}{c|c|c|c|c|c|c|c|c|c|c|c}
        \toprule
        \multirow{2}{*}{Dataset} & \multirow{2}{*}{Frames} &  \multicolumn{4}{c|}{InceptionV1} & \multicolumn{6}{c}{ResNet50} \\ 
        \cmidrule{3-12}
        & & None & I3D & Conv. &  TAM & None & I3D & Conv. &  TAM &   TSM & NLN \\ 
        \midrule
        \multirow{2}{*}{\mSSV2} & $f$=8  & 33.1 & 56.4 &  58.2 & \textbf{59.7} & 33.9 & 62.6 & 61.6 & \textbf{65.4} & 64.1 & 53.0 \\
         & $f$=16 & 34.7 & 61.8 & 63.7 & \textbf{63.9} & 35.3 & 66.2 & 65.7 & \textbf{68.6} & 67.4 & 55.0 \\
        \midrule
        \multirow{2}{*}{\mKinetics} & $f$=8  & \textbf{70.4} & 68.1 & 68.3 & 68.8 & 72.1 & 73.3 & 71.5 & \textbf{74.1} & \textbf{74.1} & 73.7 \\
         & $f$=16 & 70.5 & 70.9 & \textbf{70.7} & 70.0 & 72.5 & 75.5 & 73.4 & \textbf{76.4} & 75.6 & 74.5 \\
        \midrule
        FLOPs (G) & $f$=8 & 12.0 & 33.6 & 32.2 & 12.0 & 32.7 & 64.2 & 107.0 & 32.8 & 32.7 & 196\\
        Parameters (M) & $f$=8 & 5.7 & 12.4 & 17.2 & 5.7 & 23.7 & 46.3 & 71.5 & 23.7 & 23.7 & 31.0  \\
        \bottomrule
    \end{tabular}
    \end{adjustbox}
    \caption{Performance of different temporal aggregation strategies w/o temporal pooling. FLOPs of 16-frame are the double of 8-frame models, and number of parameters are the same.}
    \label{table:diff_ta_d} 
\end{table*}

\section{Datasets}
\label{sec:supp:datasets}
Table~\ref{table:datasets} illustrates the characteristics of the datasets used in the paper.
The \SSV2 dataset contains a total of 192K videos of 174 human-object interactions, captured in a simple setup without much background information. It has been shown that temporal reasoning is essential for recognition on this dataset~\cite{TRN:zhou2018temporal}. \Kinetics has been the most popular benchmark for deep-learning-based action approaches. It consists of 240K training videos and 20K validation videos of 400 action categories, with each video lasting 6-10 seconds. Interestingly, approaches without temporal modeling such as TSN~\cite{TSN:wang2016temporal} achieves strong results on this dataset, implying that modeling temporal information is not that important on this dataset. 
\Moments is a recent collection of one million labeled videos, involving actions from people, animals, objects or natural phenomena. It has 339 classes and each clip is trimmed to 3 seconds long. These  datasets cover a wide range of different types of videos, hence are suitable for studying various spatio-temporal representations.

We extract video frames via the FFMPEG packages and then resize the shorter side of an image to 256 while keeping the aspect ratio of the image.


\section{Training and Evaluation}
\label{sec:supp:training}
\subsection{Mini-Datasets}
\noindent \textbf{Training.}
Table~\ref{table:training_protocol} illustrates the training protocol we use for all the models in our experiments. 
We train most of our models using a single compute node with 6 V100 GPUs and a total of 96G GPU memory with a batch size of 72 or the maximum allowed for a single node (a multiple of 6). For some of the large models (for example, I3D-ResNet50) using 32 or 64 frames, we limit the number of nodes to no more than 3, i.e. 18 GPUs, and apply synchronized batch normalization in training at a batch size of 36. 
We observe that such a setup generally leads to comparable model accuracy to the approaches studied in this work. 
Following the practice in TSN~\cite{TSN:wang2016temporal}, we apply multi-scale augmentation and randomly crop the same 224$\times$224 region of whole input images for training. In the meanwhile, temporal jittering is used to sample different frames from a video. Afterward, the input is normalized by the mean and standard deviation used in the original ImageNet-pretrained model.

\begin{table}[tb]
    \centering
    \begin{adjustbox}{max width=\linewidth}
    \begin{tabular}{c|c|c|c|c}
        \toprule
            \multirow{2}{*}{Dataset} & \multicolumn{2}{c|}{\# of Images} & \# of & \multirow{2}{*}{Duration}  \\
                    &   Train & Val & Classes & \\
            \midrule
            \SSV2~\cite{Something:goyal2017something} & 168k & 24k & 174 & \multirow{2}{*}{3-5s@12fps} \\
            \mSSV2 & 81k & 12k & 87 & \\
            \midrule
            \Kinetics~\cite{Kinetics:kay2017kinetics}  & 240k & 19k & 400 & \multirow{2}{*}{6-10s@30fps} \\
            \mKinetics            & 121k & 10k & 200 & \\
            \midrule
            \Moments~\cite{Moments:monfort2019moments}        & 802k & 34k & 339 & \multirow{2}{*}{3s@30fps} \\
            \mMoments        & 100k & 10k & 200 & \\
            
        \bottomrule
        \multicolumn{5}{l}{Mini-Something-Something and Mini-Kinetics400 are.} \\
        \multicolumn{5}{l}{created by randomly sampling half of classes.} \\
    \end{tabular}
    \end{adjustbox}
    \caption{Overview of datasets.}
    \label{table:datasets}
\end{table}

\vspace{1mm}
\noindent \textbf{Evaluation.}
In the \textit{clip-level} accuracy setting, we sample $f$ frames either with uniform sampling or dense sampling and then crop a 224$\times$224 region centered at each image after resizing the shorter side of the image to 224. For uniforming sampling, we choose the middle frame of each segment to form a clip while for dense sample the first clip is used.
In the \textit{video-level} accuracy setting, $m$ clips need to be prepared. For dense sampling, we uniformly select $m$ points and then take $f$ consecutive frames starting at each point. In the case of uniform sampling, we apply an offset $i$ from the middle frame, where $-m/2 <= i < m/2$, to shift the sampling location at each segment. We use $m=10$ to compute video-level accuracy in our analysis.

\begin{figure*}[tb!]
    \centering
    \includegraphics[width=\linewidth]{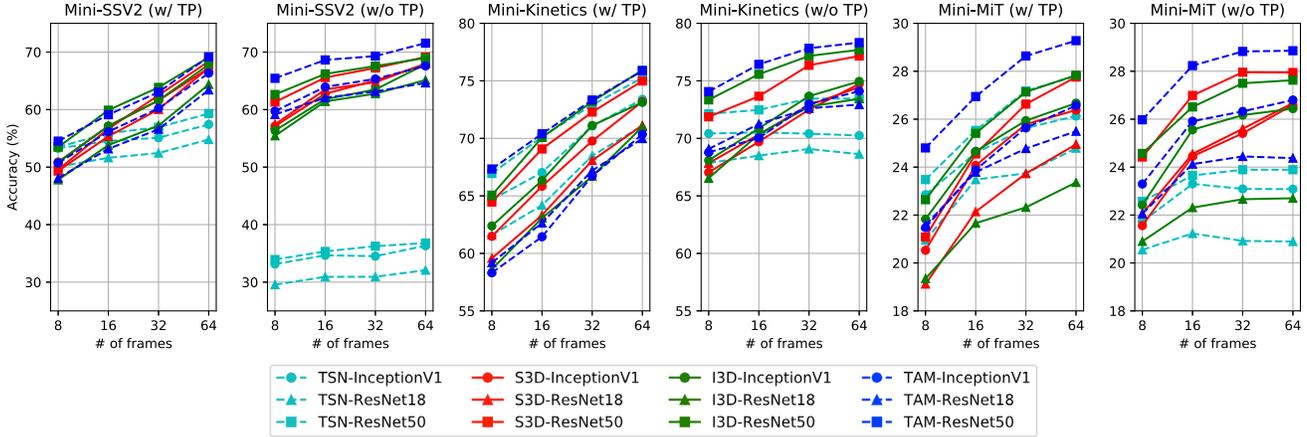}
    \caption{Top-1 accuracy of all the models with and without temporal pooling on three mini-datasets. The video architectures are separated by color while the backbones are separated by symbols. Best viewed in color.}
    \label{fig:supp:all_results_wtp}
\end{figure*}


\begin{table}[tb]
\begin{adjustbox}{max width=\linewidth}
\centering
    
        \begin{tabular}{c|cccc}
        \toprule
                              & 8-frame      & 16-frame                         & 32-frame                         & 64-frame                         \\ 
        \midrule
        Weight Init.            & ImageNet & 8-frame                       & 16-frame                     & 32-frame                       \\
        Epochs$^1$                & 75 (100) & 35 (45)                      & 35 (45)                      & 35 (45) \\
        Learning rate         & \multicolumn{4}{c}{0.01}  \\
        LR scheduler$^2$          & cosine   & multisteps & multisteps & multisteps \\
        Weight decay        &    \multicolumn{4}{c}{0.0005} \\
        Optimizer           &   \multicolumn{4}{c}{Synchronized SGD with moment 0.9} \\
        
        \bottomrule
        \multicolumn{5}{l}{\footnotesize{$^1$: Mini-Kinetics uses the epoch numbers mentioned in the bracket.}} \\
        \multicolumn{5}{l}{\footnotesize{$^2$: when the epoch number is 35, the learning rate drops 10$\times$ at the 10-th, 20-th, 30-th epoch;}} \\
        \multicolumn{5}{l}{\footnotesize{while drops 10$\times$ at the 15-th, 30-th, 40-th epoch when 45 epochs is used.}} \\
        \end{tabular}
    \end{adjustbox}
    \caption{Training protocol for Mini-datasets.}
    \label{table:training_protocol}
\end{table}

\subsection{Full Datasets}
\label{subsec:SOTA_dataset}
To enable SOTA results on \Kinetics, we follow the practices in the SlowFast paper~\cite{SlowFast:feichtenhofer2018slowfast} to train the models. During training, we take 64 consecutive frames from a video and sample every other frame as the input, i.e., 32 frames are fed to the model. The shorter side of a video is randomly resized to the range of [256, 320] while keeping aspect ratio, and then we randomly crop a 224$\times$224 spatial region as the training input. We trained all models for 196 epochs, using a total batch size of 1024 with 128 GPUs, i.e. 8 samples per GPU.  Batch normalization is computed on those 8 samples. We warm up the learning rate from 0.01 to 1.6 with 34 epochs linearly and then apply half-period cosine annealing schedule for the remaining epochs. We use synchronized SGD with momentum 0.9 and weight decay 0.0001.
During the evaluation, we uniformly sample 10 clips from a video, and then take 3 256$\times$256 crops from each clip whose shorter side of each clip is resized 256. The accuracy of a video is conducted by averaging over 30 predictions. On the other hand, for \SSV2 dataset, we only sample 2 clips since the video length of \SSV2 is shorter.

\subsection{Transfer Learning}
In transfer learning study, we used the models described in Section 6.2, i.e., 32-frame models trained with \Kinetics, as the pretrained weights. Then, we finetuned 45 epochs with cosine annealing learning rate schedule starting with 0.01, and trained the models with a batch size of 48 with synchronized batch normalization on a 6 GPUs machine.

\section{Model Implementation}
\label{sec:supp:implementation}
\noindent \textbf{Implementation.}
We follow the original published papers as much as we can to implement the approaches in our analysis. However, due to the differences in the backbones, some modifications are necessary to ensure a fair comparison under a common experimental framework. 
Here we describe how we build the networks including three backbones (InceptionV1, ResNet18 and ResNet50), four video architectures (I3D, S3D, TAM and TSN), and where to perform temporal pooling.
For three backbones, we used those 2D models available on the torchvision repository (\textit{googlenet, resnet18, resnet50}), and then used the weights in the model zoo for initializing the models either through inflation (I3D and S3D) or directly loading (TAM and TSN). Note that, for inflation, we simply copy the weights along the time dimension.
Moreover, we always perform the same number of temporal pooling at the similar locations across all the backbones while training models with temporal pooling. For each backbone, there are five positions to perform \textit{spatial pooling}, we add maximum \textit{temporal pooling} along with the last three spatial poolings (kernel size is set to 3).

\vspace{1mm}
\noindent
\textbf{I3D.} We follow I3D paper to re-implement the network~\cite{I3D:carreira2017quo}. We convert all 2D convolutional layer into 3D convolutions and set kernel size in temporal domain to 3 while using same spatial kernel size. For I3D-ResNet-50, we convert 3$\times$3 convolution in bottleneck block into 3$\times$3$\times$3.

\vspace{1mm}
\noindent
\textbf{S3D.} We follow the idea of the original S3D and R(2+1)D paper to factorize 3D convolution in the re-implemented models~\cite{S3D,R2plus1D:Tran_2018_CVPR}; thus, each 3D convolution in I3D becomes one 2D spatial convolution and one 1D temporal convolution. Nonetheless, the first convolution of the network is not factorized as the original papers. For InceptionV1 backbone, the difference from the original paper is the location of temporal pooling of backbone~\cite{S3D}. More specifically, in our implementation, we remove the temporal stride in the first convolutional layer and then add an temporal pooling layer to keep the same temporal downsampling ratio over the model. On the other hand, for ResNet backbone, 
we do not follow the R(2+1)D paper to expand the channels to have similar parameters to the corresponding I3D models, 
we simply set the output channels to the original output channel size~\cite{R2plus1D:Tran_2018_CVPR} which helps us to directly load the ImageNet-pretrained weights into the model.

\vspace{1mm}
\noindent
\textbf{TAM.} We follow the original paper to build TAM-ResNet~\cite{bLVNetTAM}, the TAM module is inserted at the non-identity path of every residual block. For TAM-InceptionV1, we add TAM modules after the every inception module.

\vspace{1mm}
\noindent
\textbf{TSN.} It does not have any temporal modeling, so it directly uses 2D models.

\end{document}